\documentclass[10pt,twocolumn,letterpaper]{article}

\PassOptionsToPackage{table}{xcolor}
\usepackage{cvpr}              

\usepackage{multirow} 
\usepackage{booktabs} 
\usepackage{natbib} 
\usepackage[T1]{fontenc}

\definecolor{cvprblue}{rgb}{0.21,0.49,0.74}
\usepackage[pagebackref,breaklinks,colorlinks,allcolors=cvprblue]{hyperref}

\def\paperID{52} 
\def\confName{EarthVision}
\def\confYear{2025}

\title{FrogDogNet: Fourier frequency Retained visual prompt Output Guidance for Domain Generalization of CLIP in Remote Sensing}

\author{Hariseetharam Gunduboina$^{1}$ \and Muhammad Haris Khan$^{2}$ \and \and Biplab Banerjee$^{1}$
\and
$^{1}$Indian Institute of Technology Bombay, India\and
$^{2}$Mohamed Bin Zayed University of Artificial Intelligence, UAE
\and
\tt\small hariseetharam552@gmail.com, muhammad.haris@mbzuai.ac.ae, getbiplab@gmail.com}

\begin{document}
\maketitle
\begin{abstract}
In recent years, large-scale vision-language models (VLMs) like CLIP have gained attention for their zero-shot inference using instructional text prompts. While these models excel in general computer vision, their potential for domain generalization in remote sensing (RS) remains underexplored. Existing approaches enhance prompt learning by generating visual prompt tokens but rely on full-image features, introducing noise and background artifacts that vary within a class, causing misclassification. To address this, we propose \texttt{FrogDogNet}, a novel prompt learning framework integrating Fourier frequency filtering and self-attention to improve RS scene classification and domain generalization. FrogDogNet selectively retains invariant low-frequency components while eliminating noise and irrelevant backgrounds, ensuring robust feature representation across domains. The model first extracts significant features via projection and self-attention, then applies frequency-based filtering to preserve essential structural information for prompt learning. Extensive experiments on four RS datasets and three domain generalization tasks show that FrogDogNet consistently outperforms state-of-the-art prompt learning methods, demonstrating superior adaptability across domain shifts. Our findings highlight the effectiveness of frequency-based invariant feature retention in generalization, paving the way for broader applications. Our code is available at \url{https://github.com/HariseetharamG/FrogDogNet}
\end{abstract}    
\section{Introduction}
\label{sec:intro}
\begin{figure}
    \centering
    \includegraphics[scale=0.078]{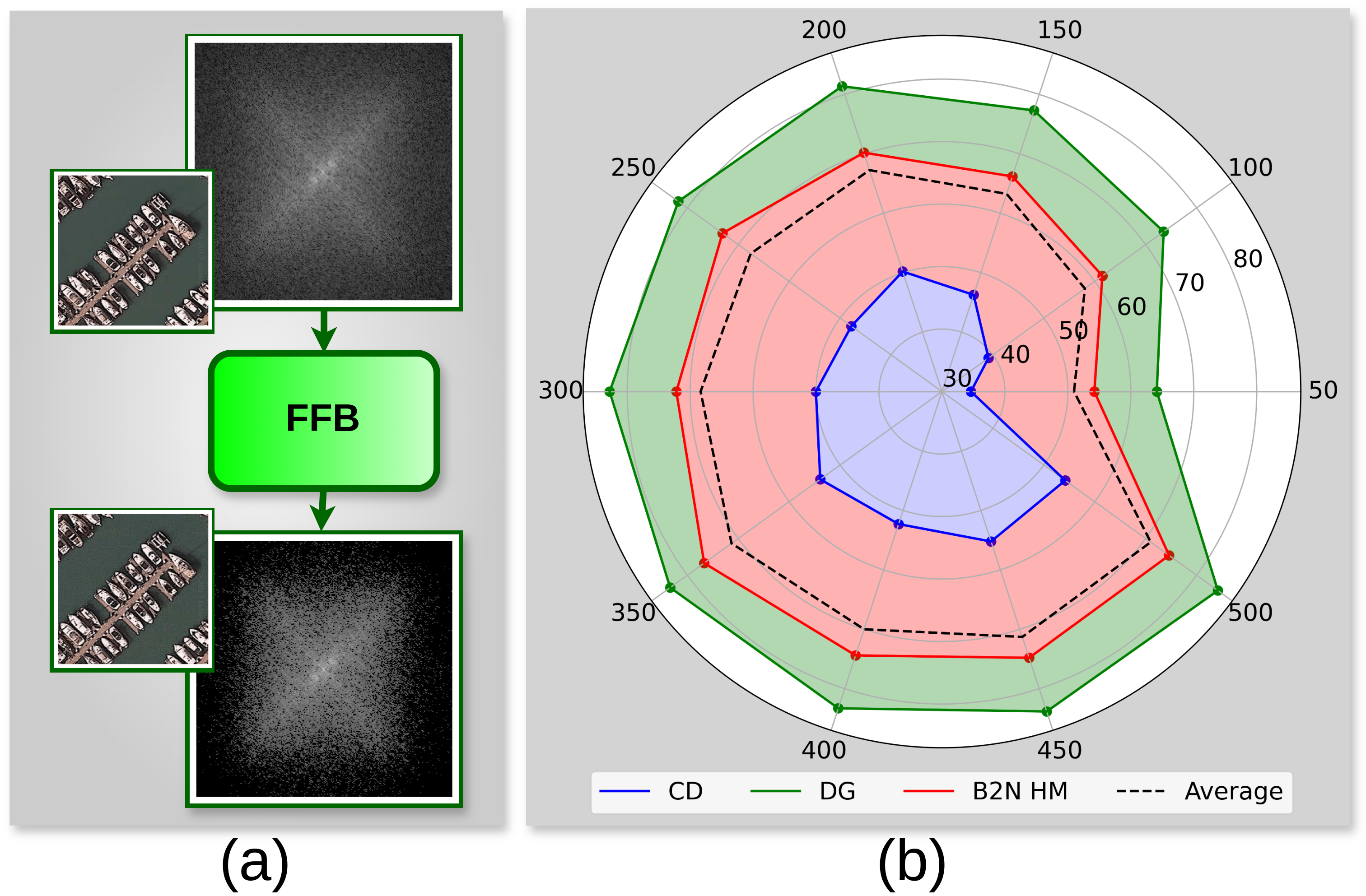}
    \vspace{-0.7cm}
    \caption{\small{Impact of Fourier Filtering on RS Image Analysis: (a) demonstrates that retaining $50\%$ of low-frequency components (LFCs) preserves structural details while reducing noise in the frequency magnitude spectrum. (b) presents a sensitivity analysis, showing that keeping 350 out of 512 LFCs of visual features \( f_v(x) \) achieves the highest average generalization performance.}}
    \label{teaser}
    \vspace{-0.8cm}
\end{figure}
Remote Sensing (RS) image analysis underpins a diverse range of tasks including Earth observation \cite{data2024multimodal}, environmental monitoring \cite{sabins1999remote, hino2018machine}, urban planning \cite{li2024deep}, and disaster management \cite{inglada2004real, van2000remote}, to name a few. Although deep learning models have substantially advanced RS analytics \cite{ma2019deep}, they often exhibit performance drops under domain shifts—variations in sensor modalities, atmospheric conditions, and geographical characteristics that cause discrepancies between training (source) and testing (target) data distributions. Two core strategies address such shifts: domain adaptation (DA), which aligns source and target distributions during training \cite{tuia2016domain, ganin2015unsupervised, farahani2021brief, saha2022multitarget}, and domain generalization (DG), which strives for robust performance in unseen target domains \cite{10223771, li2018learning, rsdg1, zhou2022domain}. DG is crucial in RS due to the difficulty of collecting labeled target-domain data. Challenges include cross-sensor variability, seasonal and atmospheric changes, geographic distribution shifts, and limited real-time data for disaster monitoring \cite{tuia2016domain, Zhu2017, morley2018integrating, hu2021generalization}.
Despite progress via adversarial training \cite{Martini2021}, meta-learning \cite{10654368}, and domain-invariant representations \cite{10387074, Ghanbarzadeh2024}, existing DG approaches struggle to handle the extreme spectral and spatial variations in RS datasets.
Beyond domain shifts, RS tasks typically suffer from limited annotated data, as labeling requires domain expertise (e.g., differentiating land-use categories or vegetation types). The substantial cost and logistical complexity of labeling across diverse environments hamper large-scale data curation, exacerbating the challenge of building robust models. Few-shot learning (FSL) \cite{russwurm2020meta, zhang2021few, ji2022few, maml, chen2019meta} partially addresses this issue by enabling models to learn from minimal supervision. Although FSL has been explored for multispectral and hyperspectral RS imagery \cite{s3net,FSL_RS1,liu2018deep,spn}, most methods remain image-centric, inadvertently capturing extensive background clutter in RS scenes, such as oceans, deserts, or forests. Since key objects of interest (e.g., urban structures, agricultural plots) are often small and spatially scattered, Standard FSL pipelines optimized for natural images often generalize poorly when applied to RS datasets. This is validated in our work, where methods like CLIP \cite{clip}, CoOp \cite{coop} and CoCoOp \cite{cocoop} struggle with RS data. 

Vision-Language Models (VLMs) like CLIP \cite{clip}, ALIGN \cite{pmlr-v139-jia21b}, Florence \cite{yuan2021florence}, and LiT \cite{zhai2022lit} leverage textual supervision for powerful zero-shot adaptation. Adaptations for RS (e.g., RS-CLIP \cite{Li2023}, RemoteCLIP \cite{Liu2024}, and ChangeCLIP \cite{dong2024changeclip}) illustrate the promise of combining visual and textual cues but remain challenged by large homogeneous backgrounds and small, high-value objects in RS. Although prompt-learning techniques \cite{coop, cocoop, prograd, clip-adapter, maple2023, stylip2024} have been proposed to refine VLMs for downstream tasks, most still operate on entire images, rendering them susceptible to overfitting on dominant yet irrelevant features.
Multi-scale methods \cite{applenet, csawbhattacharya2023c} address some of these challenges but often lack a robust mechanism to filter out noise and irrelevant background artifacts. This limitation reduces their effectiveness, particularly under varying spectral conditions where such artifacts can significantly degrade performance.

To address these gaps, we propose leveraging Fourier Transform-based filtering to enhance DG for RS image analysis. Fourier filtering is widely employed in RS for denoising and spectral feature extraction, as it separates low-frequency semantic structures from high-frequency noise and sensor artifacts. This property reduces sensitivity to environmental variations and promotes stable feature extraction. However, filtering out high-frequency components too aggressively can discard fine-grained details critical for object boundaries. To mitigate this, we introduce a self-attention mechanism \cite{vaswani2017attention} that selectively retains local cues while preserving robustness under domain shifts. As illustrated in Figure~\ref{teaser}(a), retaining the top $50\%$ low-frequency components (LFC) effectively removes high-frequency clutter but can risk losing valuable details. Our analysis in Figure~\ref{teaser}(b) indicates that retaining 350 out of 512 LFCs yields the best average performance across multiple RS datasets and generalization tasks. 

We introduce a \textbf{FrogDogNet} \textit{(Fourier frequency Retained visual prompt Output Guidance for Domain Generalization of CLIP in Remote Sensing)} framework for few-shot RS recognition, integrating a Fourier Filter Block (FFB) within a CLIP-based pipeline. The FFB focuses on semantically relevant structures while discarding background clutter, thereby improving domain generalization. A Prompt Alignment Loss aligns learned prompts with RS-specific initializations, bolstering transferability across diverse imaging conditions. To further refine feature representations, we design a lightweight Meta-Net that converts filtered embeddings into tokens optimized for downstream tasks. Our principal contributions are threefold:

\noindent  \textbf{- Fourier Filter Block (FFB)}: We propose a dedicated block that isolates essential low-frequency RS features, suppresses noise, and, combined with self-attention, selectively preserves crucial object boundaries.

\noindent \textbf{- Remote Sensing Prompt Alignment Loss}: We introduce a novel loss function that harmonizes learned prompts with various RS-specific text initializations, significantly enhancing zero-shot and few-shot adaptability.
    
\noindent \textbf{- Comprehensive Evaluation}: Extensive experiments on multiple RS benchmarks—including base-to-new class generalization, cross-dataset transfer, and multi-target adaptation—demonstrate the superior performance and practical value of our approach compared to existing methods.

\section{Related Works}
\label{sec:related work}
\noindent \textbf{(a) Domain Generalization:}
Deep learning models often struggle with domain shifts, where discrepancies between training and testing distributions lead to performance degradation. DG aims to develop models that generalize to unseen domains without requiring target data. DG approaches can be categorized into multi-source DG (Multi-DG) and single-source DG (Single-DG). Multi-DG methods leverage meta-learning \cite{finn2017model, li2018learning}, regularization techniques \cite{metareg, masf, hetdg1}, adversarial training \cite{li2018deep, MMD-AAE, matsuura2020domain}, and domain augmentation \cite{ddaig, mixstyle, FACT}. Other strategies include gradient-based dropout \cite{RSC}, episodic training \cite{hetdg4}, and domain-specific masking \cite{dmg}.

Single-DG, a more challenging yet practical setting, trains models on a single domain while ensuring generalization to unseen domains. Solutions include domain expansion \cite{volpi2018generalizing, sdg1, qiao2021uncertainty, sdg4, sdg2, xu2023simde}, adversarial attacks \cite{volpi2018generalizing, szegedy2013intriguing, sdg3, sdg1}, information bottleneck \cite{tishby2000information}, contrastive learning \cite{sdg4}, and uncertainty estimation \cite{qiao2021uncertainty}. Despite substantial progress in standard vision tasks, DG for RS image classification remains underexplored \cite{rsdg2, rsdg1}. Given the high spectral variability, heterogeneous textures, and geographically induced domain shifts in RS data, further research is needed to achieve improved generalization.

\noindent \textbf{(b) Vision-Language Models in RS:}
Foundation models, particularly large-scale Vision-Language Models (VLMs) \cite{radford2021learning, jia2021scaling, yuan2021florence}, have significantly advanced computer vision by learning robust multimodal representations. Models like CLIP \cite{clip} and VisualBERT \cite{li2019visualbert} have outperformed traditional vision models across various tasks but face challenges in unimodal learning for complex applications such as image captioning \cite{shottell} and visual question answering \cite{antol2015vqa}. VLMs achieve generalization by combining pre-trained textual embeddings (e.g., BERT \cite{devlin2018bert}, GPT \cite{radford2018improving}) with visual representations from CNNs or Vision Transformers \cite{dosovitskiy2020image}.\\
With the increasing availability of textual metadata, VLMs are being actively explored for RS applications \cite{tuia2021toward, li2024vision}, including image captioning \cite{shi2017can,lu2017exploring, zhang2017natural, zhang2019description, li2020multi, wang2020word, li2020truncation, zhao2021high, zia2022transforming}, text-based image retrieval \cite{abdullah2020textrs, rahhal2020deep, yuan2022remote, rahhal2022multilanguage, cheng2021deep, yuan2021remote, yuan2022exploring, rahhal2023contrasting}, and visual question answering \cite{lobry2020rsvqa, zheng2021mutual, chappuis2022language, rahhal2022open, bazi2022bi, yuan2022from}. Other applications include scene classification \cite{li2017zero, sumbul2017fine, quan2018structural, wang2021distance, li2022generative}, semantic segmentation \cite{chen2022semi, zhang2023language}, and object detection \cite{jiang2022few, zhang2023text, lu2023few}.\\
Recent works have sought to optimize VLMs for RS by creating large-scale caption datasets. RS5M \cite{zhang2023rs5m} introduced an automated caption dataset using BLIP-2 \cite{li2023blip2}, while RSGPT \cite{hu2023rsgpt} developed human-annotated RSICap, integrating GPT-based models for captioning and visual question answering. As RS datasets continue to grow, VLMs are expected to become increasingly central to RS applications, but their adaptation to the unique spectral, spatial, and textural challenges of RS data remains an open problem.\\
\noindent \textbf{(c) Prompt Learning for RS:} 
Prompt learning, initially developed in natural language processing (NLP) \cite{petroni2019language}, has gained traction in vision tasks by leveraging pre-trained language models for enhanced generalization. Automated prompt generation techniques, such as AutoPrompt \cite{shin2020autoprompt}, optimize prompts by identifying tokens that maximize gradient-based label likelihood. In the vision domain, CoOp \cite{coop} fine-tunes CLIP through learnable prompts for few-shot classification, while CoCoOp \cite{cocoop} extends this by conditioning prompts on image features to improve generalization. Similarly, CLIP-Adapter \cite{clip-adapter} fine-tunes lightweight adapters for visual and textual embeddings, and ProGrad \cite{prograd} refines prompt tuning by preserving foundational knowledge. Other works, such as TPT \cite{tpt}, introduce predictive consistency across different views of the same image, while MaPLe \cite{maple2023} leverages coupled vision and language prompts for progressive learning, thereby enhancing robustness in a variety of recognition tasks.

In RS, prompt learning has been explored for scene classification \cite{lin2025fedrsclip, lan2024efficient}, few-shot learning \cite{applenet,cao2024domain,cao2024domainq}, and segmentation \cite{wang2024metasegnet, zhang2024segclip, chen2024rsprompter}. To address complex textures and domain shifts in RS imagery, models like APPLeNet \cite{applenet}, StyLIP \cite{stylip2024}, C-SAW \cite{csawbhattacharya2023c}, and RS3Lip \cite{Jha2025RS3Lip} utilize multi-scale visual features, style-based prompts, or spatially-aware prompting.

\section{Methodology}
\label{sec:methodology}
\begin{figure*}
    \centering
    \includegraphics[scale=0.17]{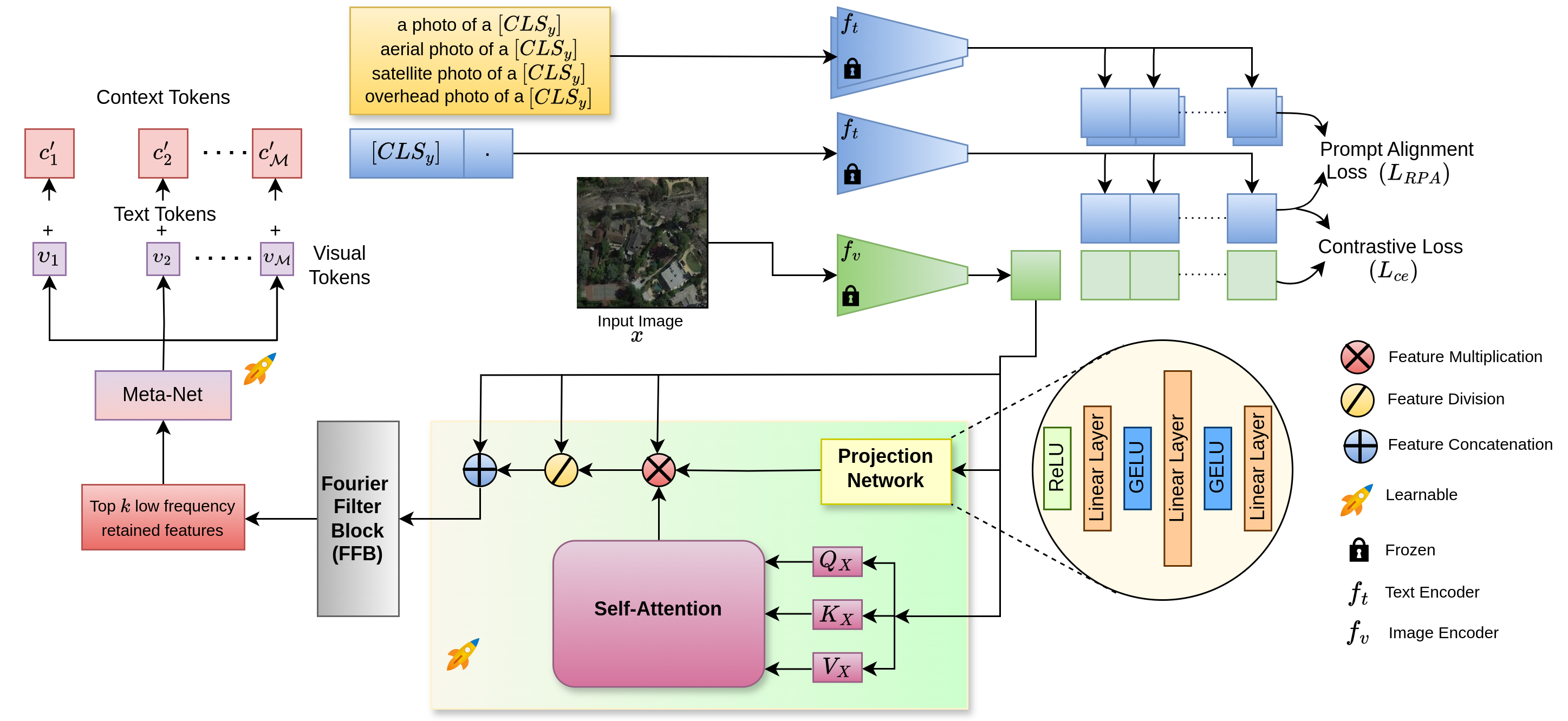}
    \vspace{-0.3cm}
    \caption{\small{Overview of FrogDogNet comprises a text encoder (\(f_t\)), an image encoder (\(f_v\)), a projection network, self-attention, and a Fourier Filter Block (FFB) for refining visual features. The image encoder extracts features, which are processed through the projection network and self-attention with residual connections, while the FFB retains key low-frequency components. The refined features pass through a lightweight Meta-Net \(\{h_m\}_{m=1}^{\mathcal{M}}\) to generate visual tokens \(\{\upsilon_m\}_{m=1}^{\mathcal{M}}\), which are combined with learnable text tokens \(\{c_m\}_{m=1}^{\mathcal{M}}\) and class embeddings before entering the text encoder. To align the learned text prompts with remote sensing (RS) prompts, we introduce RS Prompt Alignment (RPA) loss. The model is trained using a multi-task objective, incorporating both contrastive loss and RPA loss.}}
    \label{model}
    \vspace{-0.4cm}
\end{figure*}
We introduce a novel visual prompt learning method (FrogDogNet) guided by Fourier filters to enhance domain generalization in the CLIP model for remote sensing. A detailed overview of FrogDogNet approach is presented in Figure \ref{model}. To provide context for FrogDogNet's main architecture, we begin by discussing relevant baseline models, such as CLIP \cite{clip}, CoOp \cite{coop}, and CoCoOp \cite{cocoop}.
\subsection{Preliminaries}
\label{sec:baseline}  
\textbf{CLIP}~\cite{clip} is a vision-language model that aligns visual and textual representations via contrastive learning. It comprises a visual encoder $f_v$ (e.g., ResNet~\cite{he2016deep} or ViT~\cite{vit}) for input images $x$ and a text encoder $f_t$ (BERT-based~\cite{bert}) for text prompts of the form \texttt{"a photo of $[\textit{CLS}]_y$"}, where $[\textit{CLS}]_y$ denotes class $y$. Given a batch of $N$ image-caption pairs $B = \{(x_n, t_n)\}_{n=1}^{N}$, CLIP maximizes cosine similarity for the correct $N$ pairs while minimizing it for the remaining $N^2 - N$ mismatched pairs. This is achieved through a symmetric multi-class contrastive loss~\cite{sohn2016improved}:
\begin{equation}\textstyle
  L_{\text{CLIP}} = - \frac{1}{N} \sum_{n=1}^{N} \Bigl[ \log p(t_n \mid x_n) + \log p(x_n \mid t_n) \Bigr],
\end{equation}
where
\vspace{-0.2cm} 
\begin{align}\textstyle
    p(t_n \mid x_n) &= \frac{\exp(\texttt{sim}(f_v(x_n), f_t(t_n)) / \tau)}%
    {\sum_{m=1}^{N} \exp(\texttt{sim}(f_v(x_n), f_t(t_m)) / \tau)}, \\
    p(x_n \mid t_n) &= \frac{\exp(\texttt{sim}(f_t(t_n), f_v(x_n)) / \tau)}%
    {\sum_{m=1}^{N} \exp(\texttt{sim}(f_t(t_n), f_v(x_m)) / \tau)},
\end{align}
$\texttt{sim}(\cdot,\cdot)$ denotes cosine similarity and $\tau$ is the temperature parameter. By training on a large-scale dataset of 400 million image-text pairs, CLIP effectively aligns images with textual descriptions for zero-shot transfer.

Despite its robustness, CLIP relies on manually crafted prompts, limiting flexibility across domains. \textbf{CoOp}~\cite{coop} addresses this by introducing learnable prompts---context vectors optimized via backpropagation in place of static templates, structured as \(t_y = \{[c_1], [c_2], \dots, [c_{\mathcal{M}}], [\textit{CLS}_y]\}\), where \( t_y \) represents the prompt for class \( y \), \( [c_1]\dots [c_{\mathcal{M}}] \) are the \(\mathcal{M}\) learnable context tokens.
However, CoOp often struggles with domain shifts. 

\textbf{CoCoOp}~\cite{cocoop} improves generalization by conditioning prompt learning on visual features through a meta-network that refines context vectors dynamically, thereby allowing adaptation across diverse inputs. Yet, CoCoOp remains less effective in specialized fields like remote sensing and medical imaging, where unique image characteristics demand more tailored fine-tuning strategies.

\noindent\textbf{Motivation:} In RS, crucial class-defining features are frequently entangled with irrelevant backgrounds, leading to non-stationary semantics.Our FrogDogNet approach selectively propagates only the most discriminative, consistent features for each class, filtering out irrelevant information to focus on class-representative content. This targeted strategy, validated in Figure \ref{teaser}, enhances robustness and classification accuracy. In contrast, CoCoOp employs all image features indiscriminately, resulting in less effective domain transfer for RS tasks. 

To propagate only the most discriminative features into FrogDogNet's prompt learning pipeline, we leverage the \textbf{Discrete Fourier Transform (DFT)} and its computationally efficient variant, the Fast Fourier Transform (FFT) \cite{cooley1965fft}, reduce the complexity from \( O(P^2) \) to \( O(P \log P) \) \cite{rao2021global}. For a discrete sequence \(\{s_p\}\) of length \( P \), with \( p \in [0, P-1] \), the one-dimensional (1D) DFT is defined as: 
\vspace{-0.2cm}
\begin{equation}
    \displaystyle S_q = \sum_{p=0}^{P-1} s_p e^{-i\dfrac{2\pi}{P}pq}, \quad 0 \leq q \leq P-1.
    \vspace{-0.2cm}
\end{equation}  
In two-dimensional settings (e.g., images), the 1D DFT is applied along both rows and columns, forming the 2D DFT: 
\vspace{-0.2cm}
\begin{equation}
    \displaystyle S_{(q,r)} = \sum_{p=0}^{P-1} \sum_{t=0}^{T-1} s_{(p,t)} e^{-i\dfrac{2\pi}{P}pq} e^{-i\dfrac{2\pi}{T}tr},
    \vspace{-0.15cm}
\end{equation}  
where \( s_{(p,t)} \) represents the spatial domain input, \( S_{(q,r)} \) is the transformed representation, and \( P, T \) denote the respective dimensions.
By transforming images or feature embeddings into the frequency domain, we can retain only a fraction of LFC—those capturing the dominant spatial structures—while discarding high-frequency details often associated with noise or irrelevant content. The Inverse FFT (IFFT) then reconstructs the filtered images or embeddings back into the spatial domain. This approach introduces a novel advancement in visual prompt learning, addressing a limitation in earlier methods where all image features, including noise, were used in the optimization process, leading to suboptimal performance. In contrast, this method retains only the most significant visual features for prompt tuning, which are also redundant within the class, enabling the model to outperform all major prompting baselines.

   
\subsection{Model architecture of FrogDogNet}
\label{Model Architecture}
Now coming to the main model, as shown in the Figure \ref{model}, we passed the image feature embeddings \(f_v(x)\) to the Projection Network and Self Attention sections with residual connections, before passing to the FFB.
\subsubsection{Projection Network}  
Let us consider \( f_v(x) = X \), representing the image embeddings. Our goal is to refine these embeddings while preserving key information. Let \( X \in \mathbb{R}^{B \times d} \) be the input to the projection network, where \( B \) is the batch size and \( d \) is the feature dimension. We first apply a linear transformation:
\vspace{-0.1cm} 
\begin{equation} 
X' = W_1 X + b_1,  
\vspace{-0.1cm}
\end{equation}  
using learnable weights \( W_1 \in \mathbb{R}^{d \times d} \), where \( X' \) represents the transformed feature embeddings. A Gaussian Error Linear Unit (GELU) activation introduces smooth non-linearity, followed by an expansion step: 
\vspace{-0.1cm} 
\begin{equation}  
X'' = W_2 \,\mathrm{GELU}(X') + b_2, 
\vspace{-0.1cm} 
\end{equation}  
where \( X'' \) increases the feature dimension from \( d \) to \( 2d \) via \( W_2 \in \mathbb{R}^{d \times 2d} \). After another GELU activation, the features are projected back to \( d \) dimensions, denoted as \( X''' \):
\vspace{-0.1cm} 
\begin{equation}  
X''' = W_3 \,\mathrm{GELU}(X'') + b_3,
\vspace{-0.1cm} 
\end{equation}  
where \( W_3 \in \mathbb{R}^{2d \times d} \), and \( b_1, b_2, b_3 \) are bias terms. Finally, a ReLU activation is applied to encourage sparse activations, and its output is multiplied by \( X \) to form a residual connection. Let the final output be \( X_{\mathrm{projector}} \):  
\begin{equation}  
X_{\mathrm{projector}} = \mathrm{ReLU}(X''') \cdot X.  
\end{equation}  

\subsubsection{Proposed Self-Attention Module}  
To capture complex dependencies among visual features \( X \), we introduce a self-attention mechanism where the same input features serve as the query, key, and value: 
\vspace{-0.1cm}
\begin{equation}  
\text{Attention}(Q = X, K = X, V = X). 
\vspace{-0.2cm}
\end{equation}  
Specifically, each feature vector attends to all others using scaled dot-product attention: 
\vspace{-0.2cm}
\begin{equation}  
\textstyle X_{\mathrm{self}} = \text{Attention}(Q, K, V) = \text{Softmax}\left(\frac{QK^\mathrm{T}}{\sqrt{d}}\right)V, 
\vspace{-0.1cm}
\end{equation}  
where \( d \) denotes the feature dimension. These attention scores modulate the contribution of each value vector to the final output, enabling the model to dynamically aggregate contextual information. Additionally, we scale the outputs from the Projection Network \( (X_{\mathrm{projector}}) \) and the self-attention module \( (X_{\mathrm{self}}) \) by a tunable factor \( \lambda \) to regulate their influence within the architecture:
\vspace{-0.2cm}
\begin{equation}  
    X_{\mathrm{out}} = \lambda \cdot X_{\mathrm{projector}} \cdot X_{\mathrm{self}},  
    \label{tunable_factor}  
    \vspace{-0.1cm}
\end{equation}  
and subsequently normalize \( X_{\mathrm{out}} \) by dividing it by \( X \). We then incorporate a residual connection by adding \( X \): 
\vspace{-0.1cm}
\begin{equation}
X_{\mathrm{final}} = X_{\mathrm{out}} / X + X.
\vspace{-0.2cm}
\end{equation}  
We refer to \( X_{\mathrm{final}} \) as the \emph{Processed Image Features} (PIFs), which are passed to the FFB for further refinement.  
\subsubsection{Fourier Filter Block (FFB)}
\begin{figure}
    \centering
    \includegraphics[scale=0.145]{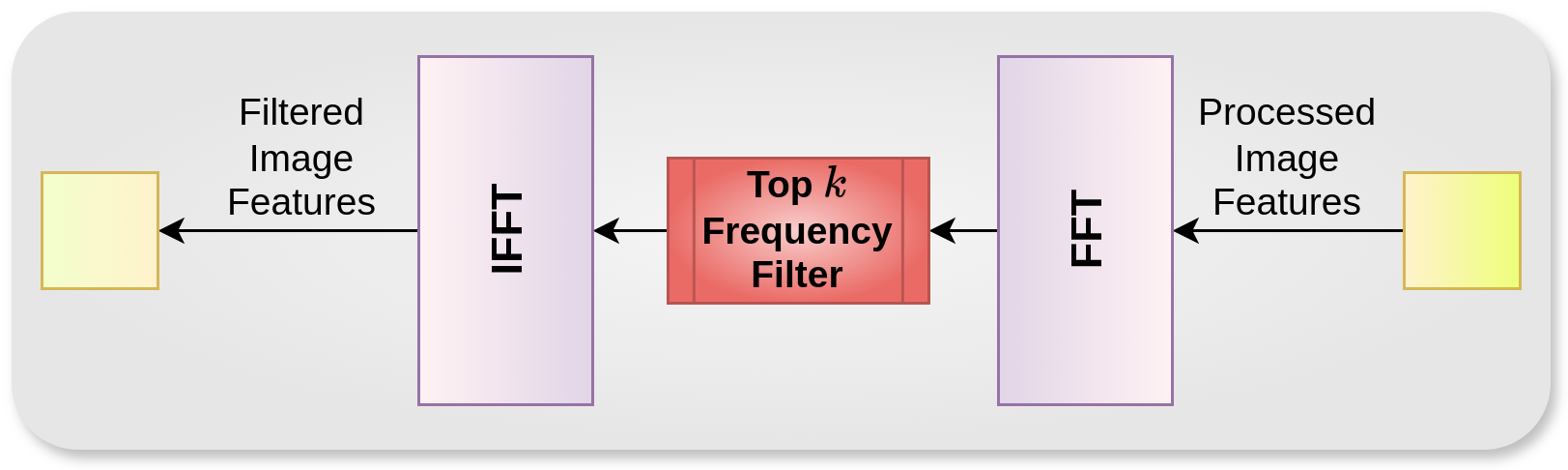}
    \vspace{-0.3cm}
    \caption{\small{The Fourier Filter Block (FFB) transforms image embeddings $X$ using Fast Fourier Transform (FFT), retains the top $k$ frequencies, and applies Inverse FFT (IFFT) to map them back to the original embedding space.}}
    \label{ffb}
    \vspace{-0.4cm}
\end{figure}
We pass the PIFs to the FFB, where they are transformed from the spatial domain to the frequency domain using FFT. We then retain top $k$ LFC, which capture the most salient yet class-redundant information in the embeddings, and apply an inverse IFFT to map the filtered features back to the spatial domain. Figure~\ref{ffb} provides an overview of this process.

After filtering, the resulting retained features $\psi \in \mathbb{R}^{B \times d}$ are fed into a lightweight neural network, termed the \emph{Meta-Net}, following the design in~\cite{cocoop}. The Meta-Net consists of a set of learnable functions $\{h_m\}_{m=1}^{\mathcal{M}}$ that generate $\mathcal{M}$ visual tokens $\{\upsilon_1, \ldots,\upsilon_{\mathcal{M}}\}$ according to:
\vspace{-0.1cm} 
\begin{equation}
\upsilon_m = h_m(\psi).
\vspace{-0.1cm} 
\end{equation}
Each visual token $\upsilon_m$ is then added to its corresponding textual token $c_m$ to form the $m^{\mathrm{th}}$ prompt token:
\vspace{-0.1cm} 
\begin{equation}
c'_m = c_m + \upsilon_m.
\vspace{-0.2cm}
\end{equation}
Collectively, these tokens compose the prompt:
\begin{equation}
t_y = \{[\upsilon_1 + c_1], \ldots, [\upsilon_{\mathcal{M}} + c_{\mathcal{M}}], [CLS_y]\}.
\end{equation}
By leveraging the most informative frequency components through the FFB and integrating them with textual embeddings via Meta-Net, this approach enhances prompt generation and yields stronger contextual representations.

\subsection{Loss functions, training, and inference}
We adopt a multi-task learning strategy with two complementary loss functions. First, we employ a \emph{supervised contrastive loss} $\mathbf{L_{ce}}$ to promote alignment between the visual feature representation $f_v(x)$ and the textual feature representation $f_t(t_y)$. This loss is realized via cross-entropy, ensuring accurate classification of seen classes.

To further enhance adaptability under various RS prompt initializations, we introduce a \emph{Remote Sensing Prompt Alignment Loss} $\mathbf{L_{RPA}}$, motivated by \cite{kgcoop23}. Formally, the label prediction for an input $x$ is given by
\begin{equation} \small
 p(y \mid x) = \frac{\exp\bigl(\texttt{sim}(f_v(x), f_t(t_y(\psi_{\phi}(x))) / \tau \bigr)}%
{\sum_{k=1}^{|\mathcal{Y}|} \exp \bigl(\texttt{sim}(f_v(x), f_t(t_k(\psi_{\phi}(x))) / \tau\bigr)},
\end{equation}
where $\phi$ indicate the learnable parameters upto FFB. 

Let \(\mathcal{D}_s = \{\mathcal{D}_s^i\}_{i=1}^{n}\) be \(n\) source domains with inputs \(x^i \in \mathcal{X}^i\) and labels \(y^i \in \mathcal{Y}_{Seen}\), where \(P(\mathcal{D}_s^i)\) varies. Training uses \(\mathcal{Y}_{Seen}\), while testing is on a target domain \(\mathcal{D}_t\) with unseen labels \(\mathcal{Y}_{Unseen}\) and \(P(\mathcal{D}_t) \neq P(\mathcal{D}_s^i)\).\\
The cross-entropy loss $\mathbf{L_{ce}}$ is defined as
\begin{equation} \small
\textstyle \mathbf{L_{ce}} = \underset{\psi_{\phi}, \{h_m\}}{\arg\min} 
\;\underset{(x,y) \in \mathcal{P}(\mathcal{D}_s)}{\mathbb{E}}
\Bigl[- \sum_{k=1}^{\mathcal{Y}_{\text{Seen}}} y_{k} \log\bigl(p(y_k \mid x)\bigr)\Bigr].
\end{equation}
We denote the textual embeddings from CLIP and from our model as $c_i^{clip} = f_t(t_i^{clip})$ and $c_i = f_t(t_i)$, respectively. Minimizing the Euclidean distance between $c_i$ and $c_i^{clip}$ strengthens alignment with RS-specific prompts, thereby improving generalization on unseen classes. Hence, we define the \emph{Remote Sensing Prompt Alignment Loss}:
\begin{equation}\textstyle
 \mathbf{L_{RPA}} = \frac{1}{Z} \sum_{z=1}^{Z} \left( \frac{1}{N_c} \sum_{i=1}^{N_c} \left\| c_i - c_{zi}^{clip} \right\|_2^2 \right), 
 \label{lrpa}
 \vspace{-0.1cm}
\end{equation}
where $\|\cdot\|_2$ denotes the Euclidean norm, $N_c$ is the number of seen classes, and $Z$ is the total number of RS prompt variants. During inference, the model leverages these aligned prompts to enhance classification accuracy on unseen classes.
Finally, the \emph{total loss} function is:
\vspace{-0.1cm}
\begin{equation}
\mathbf{L_{total}} = \underset{\psi_{\phi}, \{h_m\}}{\arg\min}\bigl[\mathbf{L_{ce}} + \Lambda \cdot \mathbf{L_{RPA}}\bigr],
\label{tunable_hyperparameter}
\vspace{-0.1cm}
\end{equation}
where $\Lambda$ is a hyper-parameter that balances the contributions of $\mathbf{L_{ce}}$ and $\mathbf{L_{RPA}}$.
During inference, cosine similarity is calculated between the image $x_t \in \mathcal{D}_t$ and the prompt embeddings of all classes in $\mathcal{Y}_{Unseen}$. The class with the highest similarity score is assigned as the predicted label:
\vspace{-0.1cm} 
\begin{equation}
    \hat{y_t} = \underset{y \in \mathcal{Y}_{Unseen}} {\arg \max} p(y|x_t).
\end{equation}

\section{Experimental Evaluations}
 
\begin{table*}[ht!]
\centering
\scriptsize{
    \centering
    \vspace{-0.3cm}
    \scalebox{0.9}{
    \begin{tabular}{lccccccccccccccc} 
    \toprule
	
\rowcolor{gray!20}&\multicolumn{3}{c}{\textbf{PatternNet}}&\multicolumn{3}{c}{\textbf{RSICD}} &\multicolumn{3}{c}{\textbf{RESISC45}}
    &\multicolumn{3}{c}{\textbf{MLRSNet}}&\multicolumn{3}{c}{\textbf{Avg. of all}}\\
      \cmidrule(lr){2-4}\cmidrule(lr){5-7}\cmidrule(lr){8-10}\cmidrule(lr){11-13}\cmidrule(lr){14-16}
     
  \rowcolor{gray!20} \multirow{-2}{*}{\textbf{Method}} &\multicolumn{1}{c}{\textbf{Base}}&\multicolumn{1}{c}{\textbf{New}}&\multicolumn{1}{c}{\textbf{HM}}
    &\multicolumn{1}{c}{\textbf{Base}}&\multicolumn{1}{c}{\textbf{New}}&\multicolumn{1}{c}{\textbf{HM}}
    &\multicolumn{1}{c}{\textbf{Base}}&\multicolumn{1}{c}{\textbf{New}}&\multicolumn{1}{c}{\textbf{HM}}
    &\multicolumn{1}{c}{\textbf{Base}}&\multicolumn{1}{c}{\textbf{New}}&\multicolumn{1}{c}{\textbf{HM}}
    &\multicolumn{1}{c}{\textbf{Base}}&\multicolumn{1}{c}{\textbf{New}}&\multicolumn{1}{c}{\textbf{HM}}\\
    
    \midrule
    \cellcolor[gray]{0.9}CLIP \cite{clip} & 63.67 & 64.37 & 64.02 & 54.61 & 55.33 & 54.97 & 56.32 & 55.38 & 55.85 & 51.43 & 51.92 & 51.67 & 56.51 & 56.75 & 56.63 \\

    \cellcolor[gray]{0.9}CoOp \cite{coop}   & 91.62 & 62.23 & 74.12 & 92.52 & 56.08 & 69.83 & 89.04 & 55.75 & 68.57 & 75.21 & 53.64 & 62.62 & 87.10 & 56.93 & 68.85 \\

    \cellcolor[gray]{0.9}CLIP-Adapter \cite{clip-adapter} & 82.15 & 63.26 & 71.48 & 78.93 & 55.44 & 65.13 & 81.67 & 56.23 & 66.60 & 71.64 & 53.19 & 61.05 & 78.60 & 57.03 & 66.10 \\

    \cellcolor[gray]{0.9}CoCoOp \cite{cocoop}& 92.39 & 63.34 & 75.16 & 93.18 & 58.67 & 72.00 & 89.78 & 57.18 & 69.86 & 76.32 & 52.75 & 62.38 & 87.92 & 57.99 & 69.88 \\

    \cellcolor[gray]{0.9}ProGrad \cite{prograd} & 92.65 & 62.48 & 74.63 & 93.44 & 58.15 & 71.69 & 90.13 & 57.89 & 70.50 & 75.96 & 52.23 & 61.90 & 88.05 & 57.69 & 69.70 \\
    \cellcolor[gray]{0.9}MaPLe \cite{maple2023} & 94.74 & 66.12 & 77.88 & 94.91 & 60.52 & 73.91 & \textbf{91.45} & 60.82 & 73.05 & 79.06 & 54.85 & 64.59 & 90.04 & 60.57 & 72.36 \\
    \cellcolor[gray]{0.9}APPLeNet \cite{applenet} & 94.89 & 65.57 & 77.55 & 95.26 & 60.71 & 74.16 & 91.24 & 60.46 & 72.73 & 78.53 & 56.41 & 65.66 & 89.98 & 60.79 & 72.56 \\
    
    \cellcolor[gray]{0.9}StyLIP \cite{stylip2024} & 95.13 & 66.78 & 78.47 & 94.98 & 60.92 & 74.23 & 90.87 & 60.34 & 72.52 & 80.65 & 55.47 & 65.73 & 90.41 & 60.87 & 72.73 \\
    \cellcolor{Goldenrod!30}FrogDogNet (ours) & \textbf{95.50} & \textbf{77.60} & \textbf{85.63} & \textbf{95.70} & \textbf{64.10} & \textbf{76.68} & 90.60 & \textbf{65.00} & \textbf{75.69} & \textbf{84.90} & \textbf{57.50} & \textbf{68.56} & \textbf{91.67} & \textbf{66.05} & \textbf{76.64} \\ 

    \midrule
    \cellcolor{SkyBlue!30} \(\Delta\) (FrogDogNet - StyLIP) & \cellcolor{green!25} +0.37 & \cellcolor{green!25} +10.82 & \cellcolor{green!25} +7.16 & \cellcolor{green!25} +0.72 & \cellcolor{green!25} +3.18 & \cellcolor{green!25} +2.45 & \cellcolor{red!25} -0.27 & \cellcolor{green!25} +4.66 & \cellcolor{green!25} +3.17 & \cellcolor{green!25} +4.25 & \cellcolor{green!25} +2.03 & \cellcolor{green!25} +2.83 & \cellcolor{green!25} +1.26 & \cellcolor{green!25} +5.18 & \cellcolor{green!25} +3.91 \\ 

    \bottomrule
    \end{tabular}}
    \captionsetup{singlelinecheck=false}
    \caption{\small{Comparison of FrogDogNet with state-of-the-art methods for the base-to-new (B2N) class generalization task. We present the accuracy for both Base and New classes, with HM denoting the harmonic mean used to evaluate the trade-off between them. The best results are shown in \textbf{bold}. \(\Delta\) shows the performance gap with StyLIP, where green (+) indicates improvement and red (-) indicates a drop.}} \footnotemark
    \label{B2N}}
    \vspace{-0.3cm}
\end{table*}
\paragraph{Datasets:}
We experiment on four RS datasets: \textbf{PatternNet}~\cite{li2018patternnet} (38 classes, 800 images/class, \(256 \times 256\)), \textbf{RSICD}~\cite{lu2017exploring} (30 classes, 10{,}000 images, \(224 \times 224\)), \textbf{RESISC45}~\cite{cheng2017remote} (45 classes, 700 images/class, \(256 \times 256\)), and \textbf{MLRSNet}~\cite{qi2020mlrsnet}(46 classes, 109{,}161 images, \(256 \times 256\)). Following \cite{applenet}, we use their \textbf{v2} version with 16 overlapping classes. Further details are in the supplementary material.

\noindent \textbf{Architecture:}
As described in Section~\ref{Model Architecture}, we use a projection network followed by a self-attention block with 4 attention heads. The resulting PIFs pass through the FFB, where we experiment with retaining various numbers of top $k$ low-frequency features. We fix $k=350$ out of 512 image features (based on Figure~\ref{teaser}(b), detailed explanation is presented in supplementary material) and project these filtered image features $\psi$ into the light weight Meta-Net to match the dimensionality of the text embeddings.

\noindent \textbf{Training and evaluation:}
We train for 50 epochs using SGD~\cite{robbins1951stochastic} with an initial learning rate of \(2\times 10^{-3}\) and a warm-up rate of \(1\times 10^{-5}\) for the first epoch. Following Figure~\ref{heatmap}, we use ViT-B/16 as the image encoder and train with 16 shots per class, a batch size of 4, and prompt variants \(Z=4\) (shown in Figure~\ref{model}). Based on extensive experiments, we set \(\lambda=0.3\) in Equation~\ref{tunable_factor} and we set \(\Lambda=0.5\) in Equation~\ref{tunable_hyperparameter}, analysis shown in Figure~\ref{hyperparameter}. Text prompts follow the template \texttt{"a photo of a [CLS]"} with a context length of four, consistent with prior works~\cite{coop, cocoop}. We report top-1 average accuracy over three random seeds.
\subsection{Comparison to the literature}  
We compare the performance of FrogDogNet with existing state-of-the-art (SOTA) methods (explained in supplementary) for three domain generalization tasks like \cite{applenet}.\\  
\noindent\textbf{1. \textit{Base-to-New Class Generalization (B2N)}:} Evaluates the model's ability to generalize to classes not seen during training, ensuring $\mathcal{Y}_{Seen}\cap\mathcal{Y}_{Unseen}=\emptyset$.  
\textbf{2. \textit{Cross-Dataset Generalization (CD)}:} Involves training on one dataset and testing on other datasets with different domain distributions and label sets, ensuring \(\mathcal{Y}_{Seen} \cap \mathcal{Y}_{Unseen} = \emptyset\) and \(P(\mathcal{D}_s) \neq P(\mathcal{D}_t)\). 
\textbf{3. \textit{Single-Source Multi-Target Domain Generalization (DG)}:} The model, trains on a single source domain and evaluates on multiple unseen target domains in a closed-set setting, where $\mathcal{Y}_{Seen}\cap\mathcal{Y}_{Unseen} = \mathcal{Y}_{Seen}\cup\mathcal{Y}_{Unseen}$ but \(P(\mathcal{D}_s) \neq P(\mathcal{D}_t)\).  

\noindent\textbf{Base-to-New Class Generalization (B2N):}
Table \ref{B2N} summarizes the B2N class generalization results across four RS datasets, using the harmonic mean (HM) to balance accuracy between base and new classes. FrogDogNet achieves the best performance, surpassing existing methods on all datasets. Compared to CLIP’s zero-shot approach, it enhances base class accuracy by $35.16\%$ and new class accuracy by $9.30\%$ on average. It also significantly improves upon adaptation-based techniques like CoOp and CoCoOp, with $7.79\%$ and $6.76\%$ higher average HM-scores, respectively. Moreover, FrogDogNet outperforms StyLIP, with a $3.91\%$ boost in the average HM-score, highlighting its superior ability to generalize to unseen classes.
\begin{table}[!ht]
\centering
\scriptsize{
    \centering
    \vspace{-0.2cm}
    \scalebox{0.9}{
    \begin{tabular}{lcccc} 
    \toprule
   \rowcolor{gray!20} &\multicolumn{1}{c}{\textbf{Source}}&\multicolumn{3}{c}{\textbf{Target}} \\
     
    \cmidrule(lr){2-2}\cmidrule(lr){3-5}
     
  \rowcolor{gray!20}  \multirow{-2}{*}{\textbf{Method}}&\multicolumn{1}{c}{\textbf{PatternNet}}&\multicolumn{1}{c}{\textbf{RSICD}}&\multicolumn{1}{c}{\textbf{RESISC45}}
    &\multicolumn{1}{c}{\textbf{MLRSNet}}\\
    
    \midrule
    \cellcolor{gray!25}CLIP \cite{clip} & 61.72 & 43.25 & 48.56 & 45.13 \\

    \cellcolor{gray!25}CoOp \cite{coop} & 85.23 & 42.53  & 49.34 & 44.50  \\

    \cellcolor{gray!25}CLIP-Adapter \cite{clip-adapter} & 74.27& 42.57  & 49.07  & 44.17  \\

    \cellcolor{gray!25}CoCoOp \cite{cocoop} & 85.95 & 43.61 & 49.53 & 44.72\\

    \cellcolor{gray!25}ProGrad \cite{prograd} & 86.14 & 41.25 & 48.26 & 44.12 \\
    \cellcolor{gray!25} MaPLe \cite{maple2023} & 87.92 & 45.23 & 49.56 & 46.37 \\
   \cellcolor{gray!25} APPLeNet \cite{applenet} & 88.17 & 44.87 & 50.97 & 46.83 \\
   \cellcolor{gray!25} StyLIP \cite{stylip2024} & 88.01 & 46.12 & 49.89 & 46.94 \\
   \cellcolor{Goldenrod!30} FrogDogNet (ours) & \textbf{91.60} & \textbf{53.10} & \textbf{56.30} & \textbf{52.30}\\ 
    
    \midrule
    \cellcolor{SkyBlue!30} \(\Delta\) (FrogDogNet - StyLIP) 
    & \cellcolor{green!25} +3.59 & \cellcolor{green!25} +6.98 & \cellcolor{green!25} +6.41 & \cellcolor{green!25} +5.36 \\ 

    \bottomrule
    \end{tabular}}
    \caption{\small{Comparison of FrogDogNet with state-of-the-art methods for cross-dataset (CD) generalization with PatternNet dataset as the source domain and remaining RS datasets as the target domains. We use the accuracy metric as the performance measure. The best results are highlighted in \textbf{bold}.}}
    \label{CDT}}
    \vspace{-0.3cm}
\end{table}

\noindent \textbf{Cross-Dataset Generalization (CD):}
Table~\ref{CDT} compares CD generalization results, treating target domains as unseen (zero-shot). Using PatternNet as the source, FrogDogNet outperforms all baselines on RSICD, RESISC45, and MLRSNet. Notably, it improves upon CLIP’s zero-shot accuracy by $9.85\%$, $7.74\%$, and $7.17\%$, respectively, and surpasses adaptation-based approaches like CoOp, CoCoOp, and StyLIP. Compared to StyLIP, FrogDogNet achieves gains of $6.98\%$, $6.41\%$, and $5.36\%$ on RSICD, RESISC45, and MLRSNet, respectively, demonstrating superior robustness under domain and label shifts.

\noindent \textbf{Single-Source Multi-Target Domain Generalization (DG):}
Table~\ref{DG} reports DG results on \textbf{v2} datasets, where 16 identical classes span all domains. We use PatternNetv2 as the source and rest all as the targets. FrogDogNet obtains the highest accuracy on RSICDv2 ($82.50\%$), RESISC45v2 ($86.90\%$), and MLRSNetv2 ($80.70\%$). Against StyLIP, it offers improvements of $1.83\%$, $2.34\%$, and $5.04\%$ on RSICDv2, RESISC45v2, and MLRSNetv2, respectively, setting a new state of the art in DG. Although StyLIP slightly surpasses FrogDogNet on source by $0.05\%$, our sizable gains on the targets confirm robust generalization.

\begin{table}[!ht]
\centering
\scriptsize{
    \centering
    \scalebox{0.82}{
    \begin{tabular}{lcccc} 
    \toprule
   \rowcolor{gray!20} &\multicolumn{1}{c}{\textbf{Source}}&\multicolumn{3}{c}{\textbf{Target}} \\
     
    \cmidrule(lr){2-2}\cmidrule(lr){3-5}
     
   \rowcolor{gray!20}   \multirow{-2}{*}{\textbf{Method}}&\multicolumn{1}{c}{\textbf{PatternNetv2}}&\multicolumn{1}{c}{\textbf{RSICDv2}}&\multicolumn{1}{c}{\textbf{RESISC45v2}}
    &\multicolumn{1}{c}{\textbf{MLRSNetv2}}\\
    
    \midrule
  \cellcolor[gray]{0.9}  ERM \cite{erm} & 73.69  & 61.40  & 61.59 & 61.13  \\
   \cellcolor[gray]{0.9} CLIP \cite{clip} & 78.04 & 72.15 & 75.42 & 67.78 \\
    
   \cellcolor[gray]{0.9} DANN \cite{dann} & 93.56 & 75.49  & 76.18  & 70.53  \\

   \cellcolor[gray]{0.9} CoOp \cite{coop} & 94.25 & 76.50  & 77.87& 70.97  \\

    \cellcolor[gray]{0.9}CLIP-Adapter \cite{clip-adapter} & 92.36 & 79.17 & 79.76 & 71.04  \\

 \cellcolor[gray]{0.9}   CoCoOp \cite{cocoop} & 94.41 & 79.33 & 80.43 & 71.67 \\

    \cellcolor[gray]{0.9}ProGrad \cite{prograd} & 95.18 & 77.46 & 80.65 & 72.29  \\
   \cellcolor[gray]{0.9} MaPLe \cite{maple2023} & 96.52 & 80.45 & 83.37 & 76.15 \\
   \cellcolor[gray]{0.9} APPLeNet \cite{applenet} & 96.63 & 81.03 & 82.23 & 74.03 \\ 
   \cellcolor[gray]{0.9} StyLIP \cite{stylip2024} & \textbf{96.85} & 80.67 & 84.56 & 75.66 \\
   \cellcolor{Goldenrod!30} FrogDogNet (ours) & 96.80 & \textbf{82.50} & \textbf{86.90} & \textbf{80.70} \\\midrule

    \rowcolor{SkyBlue!30} \(\Delta\)  (FrogDogNet  - StyLIP) 
    & \cellcolor{red!25} -0.05 
    & \cellcolor{green!25} +1.83 
    & \cellcolor{green!25} +2.34 
    & \cellcolor{green!25} +5.04 \\ 

    \bottomrule
    \end{tabular}}
     \caption{\small{Comparing FrogDogNet with state-of-the-art methods for single-source multi-target domain generalization (DG) on \textbf{V2} RS datasets. We use the accuracy metric as the performance measure. The best results are highlighted in \textbf{bold}.}}
    \label{DG}}
    \vspace{-0.6cm}
\end{table}
\subsection{Ablation analysis}
\begin{figure}
    \centering
    \includegraphics[scale=0.237]{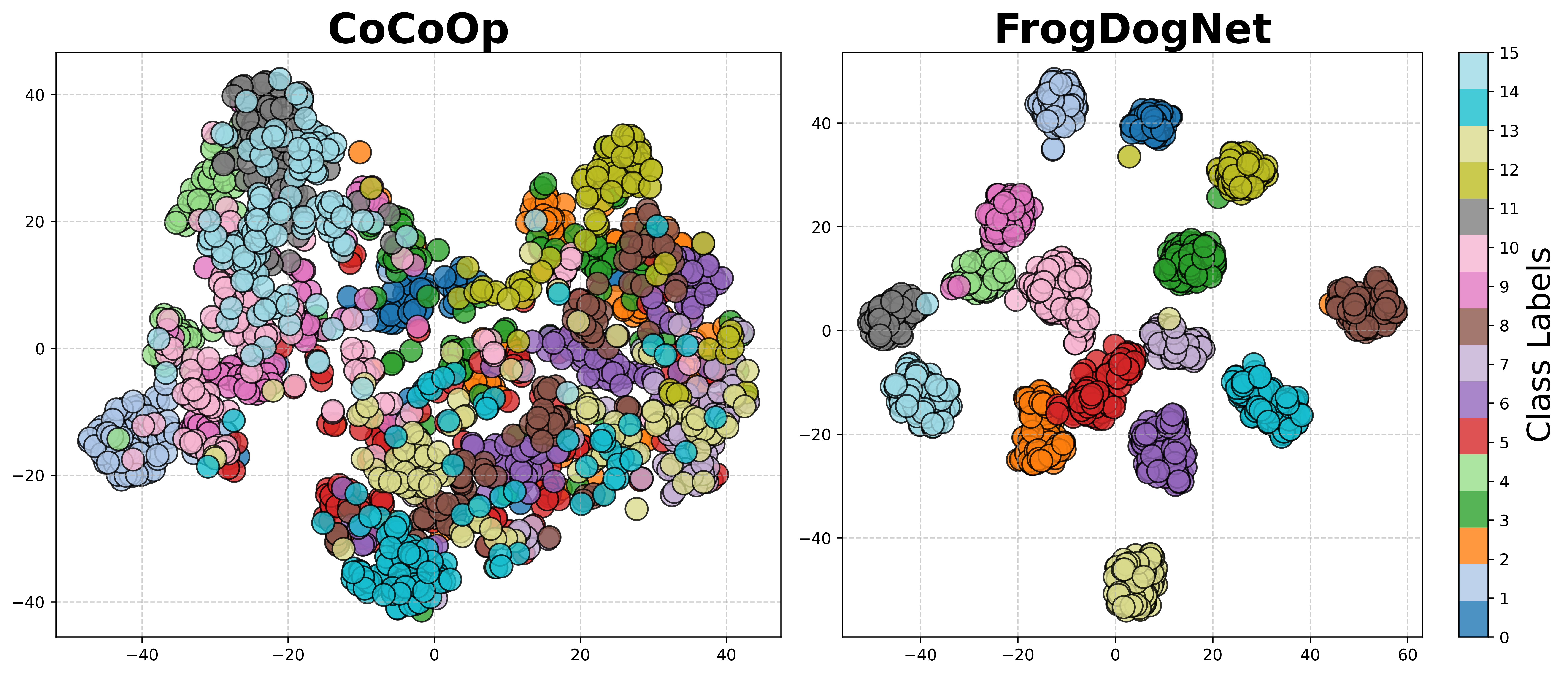}
    \vspace{-0.3cm}
    \caption{\small{t-SNE plots \cite{van2008visualizing} depicting the image features extracted from the Meta-Net of CoCoOp and the FFB of FrogDogNet for the domain generalization (DG) task on the RSICDv2 dataset. The legends indicate the corresponding class labels.}}
    \label{tsne}
    \vspace{-0.2cm}
\end{figure}

\noindent\textbf{Feature space visualization:} Figure \ref{tsne} illustrates a t-SNE \cite{tsne} plot comparing the image embeddings produced by FFB of FrogDogNet and Meta-Net of CoCoOp \cite{cocoop} on one of the target datasets RSICDv2 for the single-source multi-target domain generalization (DG) task. The results highlight FrogDogNet’s ability to form well-separated clusters for each class, whereas CoCoOp exhibits significant overlap between class clusters. This demonstrates the superior discriminability of FrogDogNet in feature space.

\noindent\textbf{Sensitivity to variation in the number of shots:} We assess the performance of FrogDogNet by varying the number of shots from 1 to 32 for the B2N class generalization task and compare it against some of the major SOTA prompting techniques, as presented in Table \ref{shots}. In this setup, we employ a context length of 4, position the class token at the end, utilize ViT-B/16 as the visual feature backbone, and adopt a unified context vector. We focus on few-shot prompting methods, reporting results on the PatternNet dataset.  
FrogDogNet consistently surpasses benchmark prompt-learning approaches, achieving at least $11\%$, $6\%$, $10.4\%$, and $0.2\%$ improvements for 4, 8, 16, and 32 shots, respectively.
\footnotetext{We do not consider comparing against \cite{csawbhattacharya2023c} and \cite{Jha2025RS3Lip} as their reliance on additional self-supervised objectives, contrastive to our model.}

\begin{figure}
    \centering
    \includegraphics[scale=0.3]{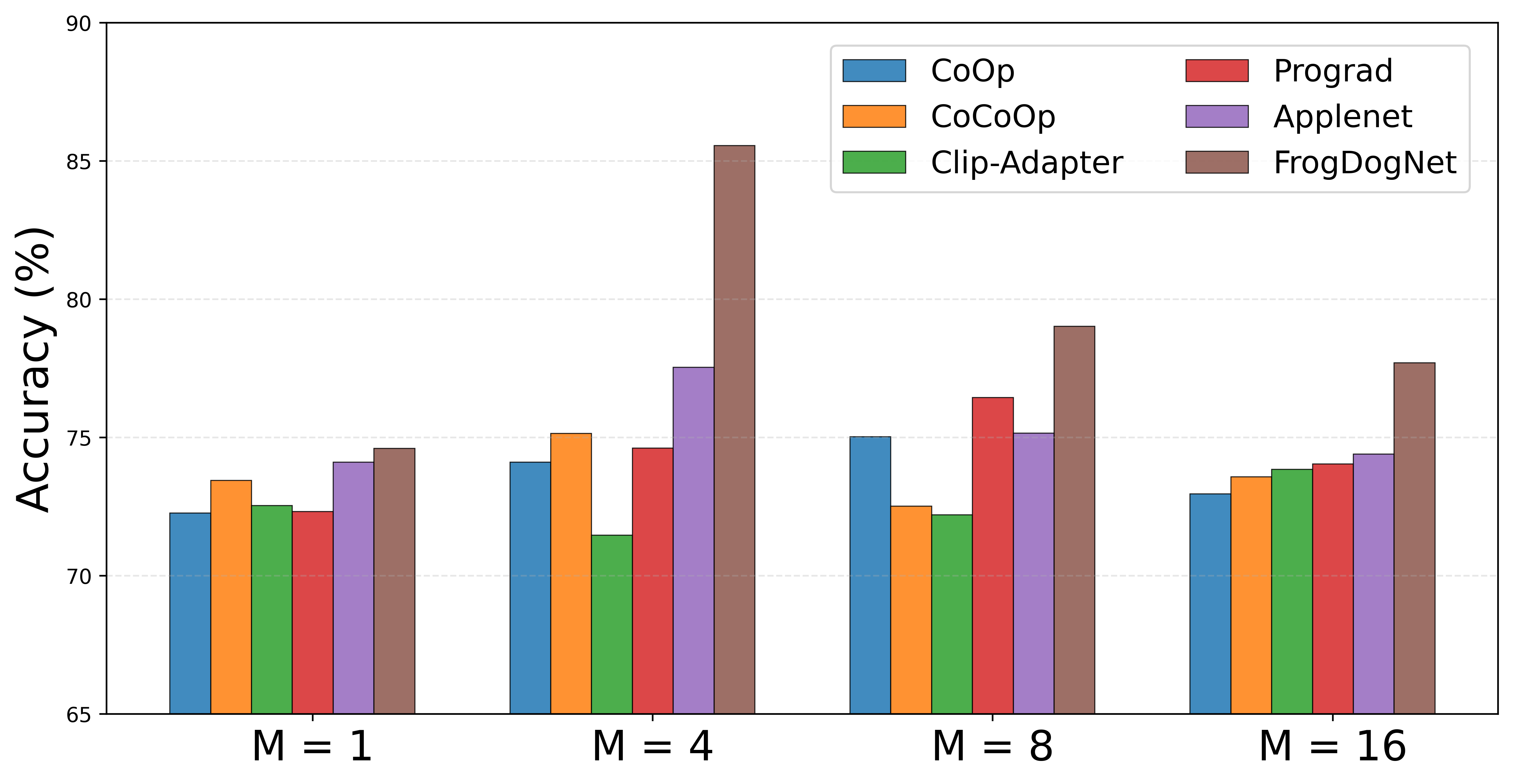}
    \vspace{-0.2cm}
    \caption{\small{Classification performance of FrogDogNet across context lengths (\(\mathcal{M}\)) for the B2N generalization task on PatternNet, compared to state-of-the-art methods, using the harmonic mean (HM) of base and new class accuracies.}}
    \label{context_length}
    \vspace{-0.4cm}
\end{figure}

\begin{table}[!ht]
 \vspace{-0.1cm}
\centering
\scriptsize{
    \centering
    \scalebox{1.0}{
    \begin{tabular}{lccccc} 
    \toprule
   \rowcolor{gray!20} \multirow{1}{*}{\textbf{Method}}&\multicolumn{1}{c}{\textbf{1-shot}}&\multicolumn{1}{c}{\textbf{4-shots}}
    &\multicolumn{1}{c}{\textbf{8-shots}}
    &\multicolumn{1}{c}{\textbf{16-shots}}
    &\multicolumn{1}{c}{\textbf{32-shots}} \\
     
     
    
    \midrule

    \cellcolor{gray!25}CoOp \cite{coop} & 70.33 & 71.61 & 72.17 & 74.12 & 74.58  \\

    \cellcolor{gray!25}CLIP-Adapter \cite{clip-adapter} & 69.75 & 69.95 & 70.37 & 71.48 & 71.64  \\

    \cellcolor{gray!25}CoCoOp \cite{cocoop} & 71.85 & 73.61 & 74.53 & 75.16 & 74.39 \\

    \cellcolor{gray!25}ProGrad \cite{prograd} & \textbf{73.67} & 72.05 & 73.16 & 74.63 & 75.56  \\

    \cellcolor{gray!25}APPLeNet \cite{applenet} & 72.44 & 72.46 & 75.28 & 77.55 &
    77.13\\ 
    \cellcolor{Goldenrod!30}FrogDogNet (ours) & 73.50 & \textbf{81.70} & \textbf{79.83} & \textbf{85.63} &
    \textbf{77.28}\\ \bottomrule
    \end{tabular}}
    \caption{\small{Comparison of FrogDogNet with state-of-the-art methods by varying the number of shots for the B2N class generalization task on the PatternNet dataset. The harmonic mean (HM) of base and new class accuracies is used to evaluate performance and illustrate the generalization trade-off. The best results are in \textbf{bold}.}}
    \label{shots}}
    \vspace{-0.6cm}
\end{table}
\noindent\textbf{Sensitivity analysis of FrogDogNet to context lengths:}  
To evaluate the impact of context length on performance of FrogDogNet, we experimented with four different values of \(\mathcal{M}\): 1, 4, 8, and 16 by initializing all \(\mathcal{M}\) length prompts in RS. As depicted in Figure \ref{context_length}, FrogDogNet outperforms all major state-of-the-art prompting methods significantly for all context lengths. Notably, FrogDogNet achieves optimal performance at \(\mathcal{M} = 4\) among all tested settings.

\begin{figure}
    \centering
    \includegraphics[scale=0.35]{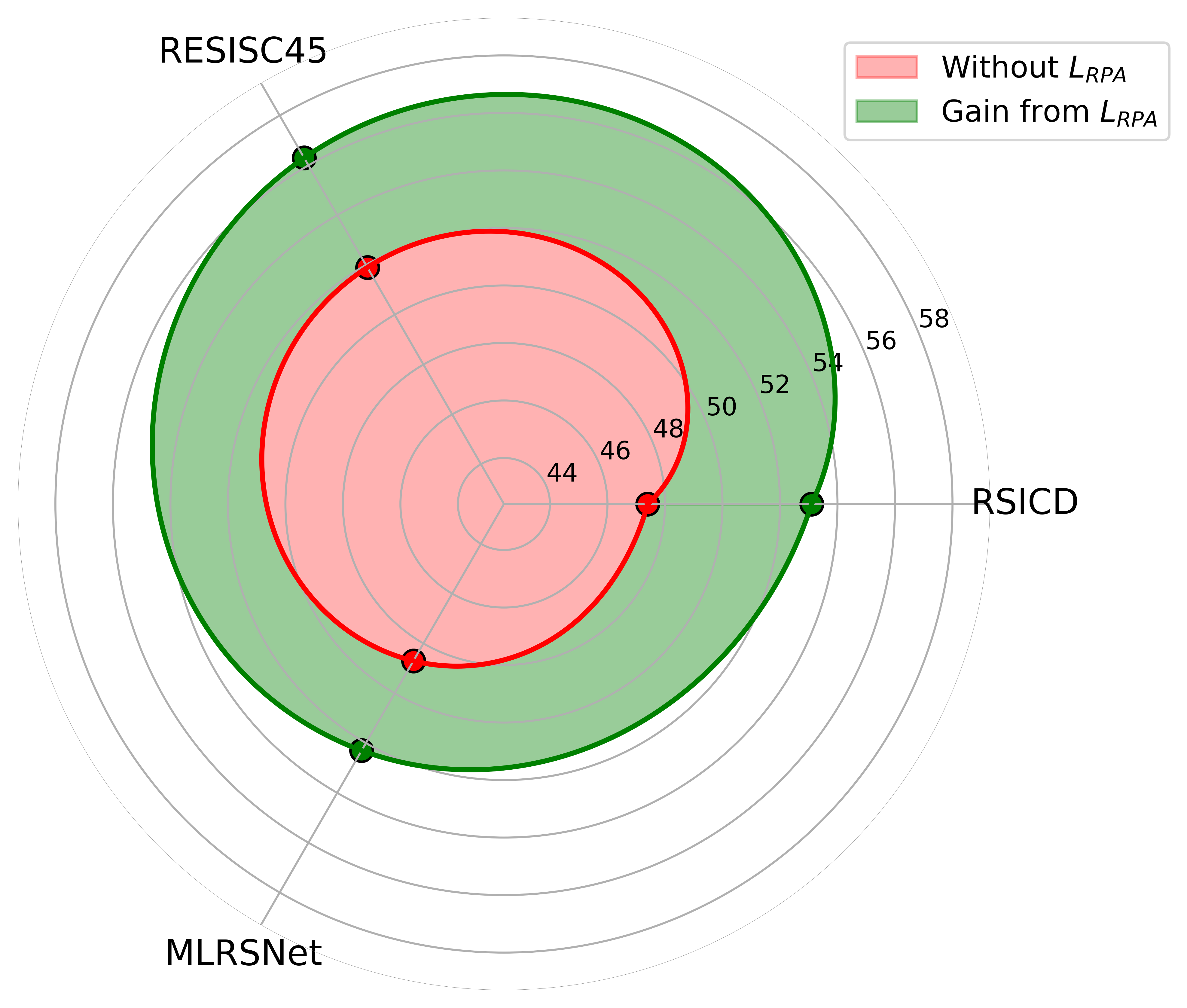}
    \caption{\small{Effect of $L_{RPA}$ on cross-dataset (CD) generalization with PatternNet as the source with prompt \texttt{"a photo of a"} and other RS datasets as the target with prompt \texttt{"satellite photo of a"}. We use accuracy as the performance measure.}}
    \label{fff_loss}
    \vspace{-0.4cm}
\end{figure}

\noindent\textbf{Effect of RPA loss (\(\mathbf{L_{RPA}}\)):}  
Figure \ref{fff_loss} presents study on the impact of RPA loss (\(\mathbf{L_{RPA}}\)) from Equation \ref{lrpa}, comparing FrogDogNet trained with and without \(\mathbf{L_{RPA}}\). Notably, FrogDogNet achieves an average performance gain of $9.3\%$ across all target domain datasets in the CD generalization task with training prompt of \texttt{"a photo of a"} but different testing prompt \texttt{"satellite photo of a"} when \(\mathbf{L_{RPA}}\) is incorporated. This highlights the crucial role of \(\mathbf{L_{RPA}}\) in enhancing feature alignment, 
allowing prompt to align better with RS prompt initializations.

\begin{figure}
    \centering
    \includegraphics[scale=0.3]{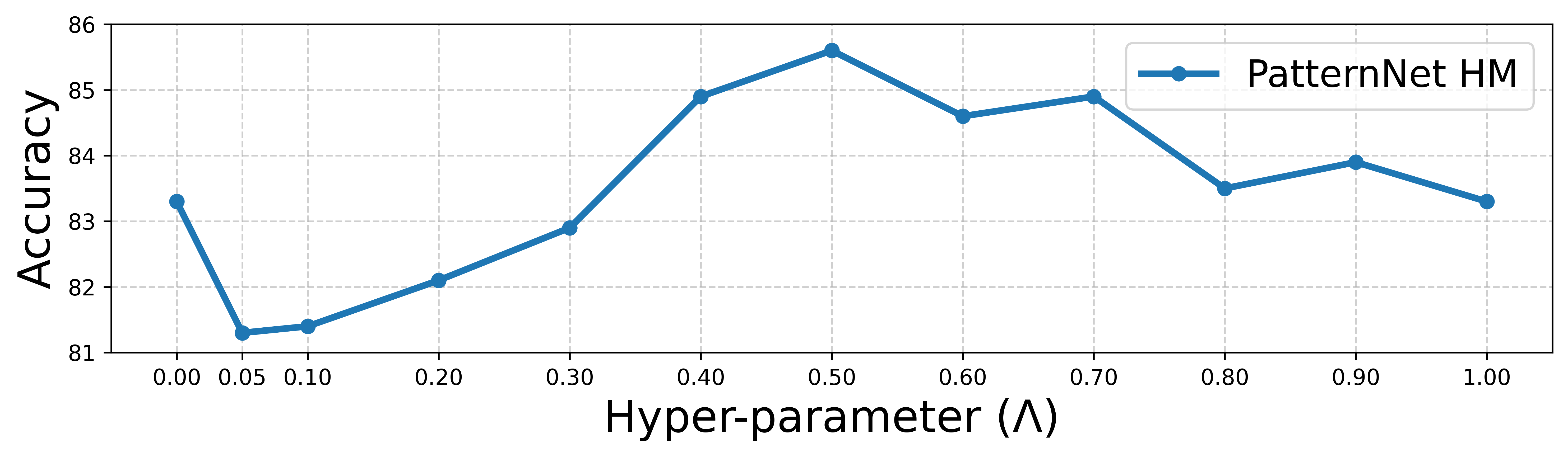}
    \vspace{-0.2cm}
    \caption{\small{Variation in FrogDogNet’s classification performance with different hyper-parameter \((\Lambda)\) values on the PatternNet dataset for the B2N generalization task, measured using the harmonic mean (HM) of base and new class accuracies.}}
    \label{hyperparameter}
    \vspace{-0.6cm}
\end{figure}

\noindent\textbf{Insights on hyper-parameter variation:} 
FrogDogNet’s performance on the PatternNet dataset fluctuates with different hyper-parameter $\Lambda$ (in equation \ref{tunable_hyperparameter}) values shown in Figure \ref{hyperparameter}. The highest HM score $85.6\%$ occurs at $0.5$, while extreme values $0.0$ and $1.0$ lead to lower performance $83.3\%$. A stable and strong performance is observed between $0.3$ and $0.7$, highlighting this range as optimal for balancing base and new class generalization. Similar trends observed in rest of the datasets (shown in supplementary).
\noindent\textbf{Prompting complexity:}
Table \ref{gfLOPs} compares the computational complexity of FrogDogNet on the PatternNet dataset with \cite{cocoop} and \cite{applenet} for the Base-to-New (B2N) task, measured using \cite{pytorchopcounter}. FrogDogNet achieves superior performance (Table \ref{B2N}) while maintaining a complexity of $192.361$ GFLOPS—$15 million$ FLOPS lower than APPLeNet and just $0.0047\%$ higher than CoCoOp, highlighting its computational efficiency despite improved performance.
\begin{table}[!ht]
\centering
\scriptsize{
    \scalebox{1.0}{
    \begin{tabular}{l c c c c} 
    \toprule
   \rowcolor{gray!20} \textbf{Method} & \textbf{CoCoOp \cite{cocoop}} & \textbf{APPLeNet \cite{applenet}} & \textbf{FrogDogNet (ours)} &  \\
    \midrule
    \cellcolor{Goldenrod!30}GFLOPS & 192.352 & 192.376 & \cellcolor{green!25}192.361  \\
    \bottomrule
    \end{tabular}
    }
}
\caption{\small{Comparison of computational complexity of FrogDogNet on the PatternNet dataset in B2N setting, where GFLOPS indicate Giga Floating Point Operations Per Second.}}
\label{gfLOPs}
\vspace{-0.3cm}
\end{table}

\begin{figure}
    \centering
    \includegraphics[scale=0.3]{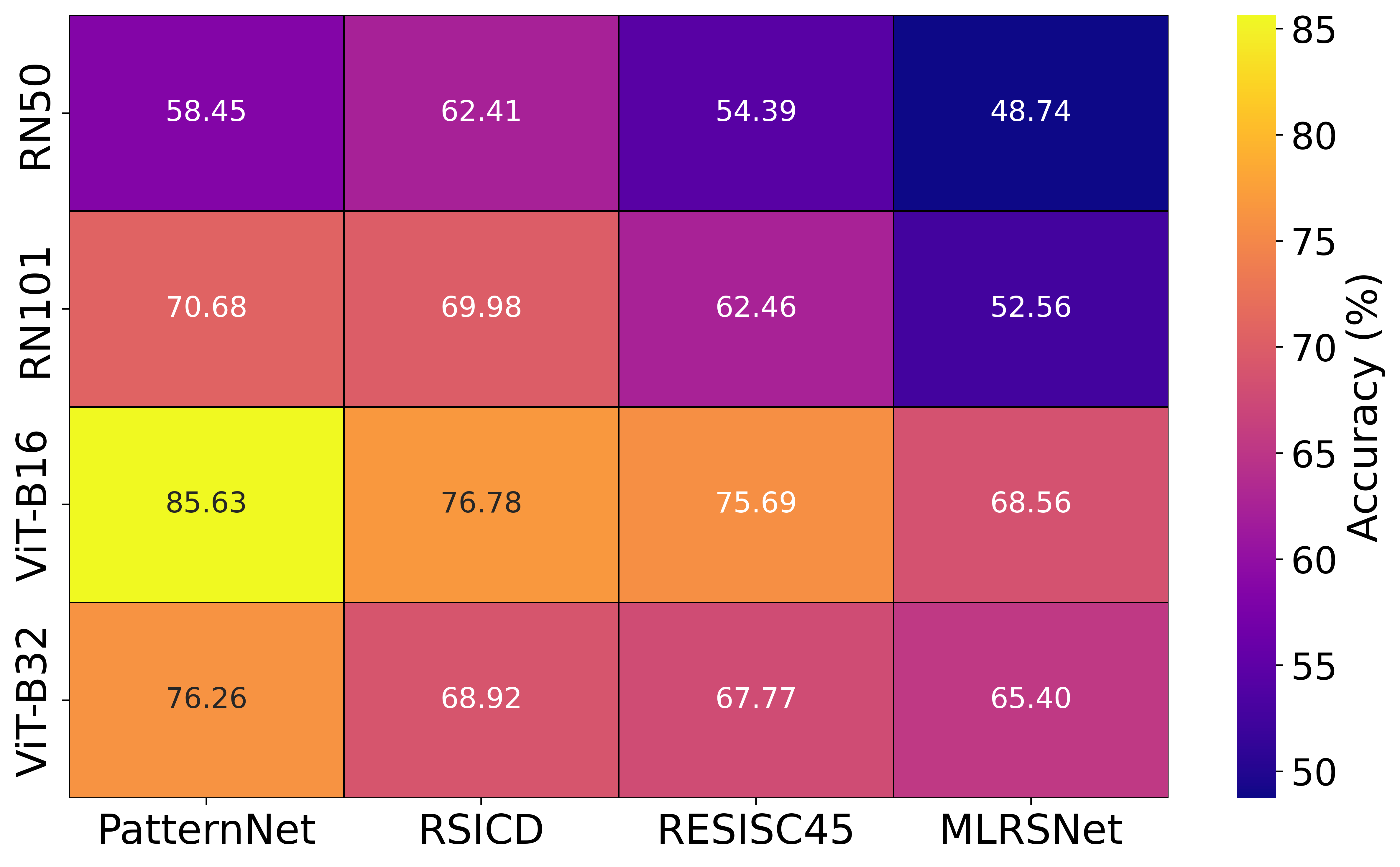}
    \vspace{-0.2cm}
    \caption{\small{Variation in FrogDogNet’s classification performance with different backbone models for the B2N class generalization task across all datasets. The harmonic mean (HM) of base and new class accuracies is used as the evaluation metric.}}
    \label{heatmap}
    \vspace{-0.4cm}
\end{figure}

\noindent\textbf{Backbone selection for FrogDogNet:}  
We evaluate FrogDogNet’s performance using different backbone models—RN50, RN101, ViT-B16, and ViT-B32—across all four datasets. As shown in Figure \ref{heatmap}, ViT-B16 consistently achieves the highest accuracy across all datasets, demonstrating its superior feature representation capabilities.
     
     
    
\section{Conclusions}
This paper introduces \textbf{FrogDogNet}, a prompt learning framework that integrates Fourier frequency filtering to improve domain generalization in remote sensing. To address CLIP’s limited generalization on remote sensing images, we introduce a novel approach that combines Fourier-based invariant feature retention with self-attention for more effective prompt learning. FrogDogNet first applies projection and self-attention to refine image features, then removes noise and background artifacts by selectively retaining essential low-frequency components before feeding them into the prompt learning process.
This strategy enables better generalization by preserving invariant features within a class, making it more robust across diverse domains. 

Extensive experiments demonstrate its superiority over existing prompt learning methods. While frequency-based filtering is widely used in image processing, its application to domain generalization remains underexplored. This work introduces a novel perspective, opening avenues for broader applications across various generalization tasks.
\clearpage

{
    \small
    \bibliographystyle{ieeenat_fullname}
    \bibliography{main}
}


\end{document}